\documentclass[10pt,twocolumn,letterpaper]{article}

\usepackage{iccv}
\usepackage{times}
\usepackage{graphicx}
\usepackage{amsmath}
\usepackage{amssymb}
\usepackage{algorithm}
\usepackage{algpseudocode}
\algtext*{EndWhile}% Remove "end while" text
\algtext*{EndIf}% Remove "end if" text
\algtext*{EndFor}% Remove "end for" text
\usepackage{epsfig}
\usepackage{booktabs}
\usepackage{multicol}
\usepackage{multirow}
\usepackage{dcolumn}
\usepackage{tabularx}
\usepackage{diagbox}
\usepackage{subfig}
\usepackage{xcolor}
\usepackage[normalem]{ulem}
\usepackage{authblk}
\usepackage{etoolbox}
\usepackage{hyperref}
\usepackage{enumitem}

\newcommand*{\figref}[2][]{%
    \hyperref[{fig:#2}]{%
        Figure~\ref*{fig:#2}%
        \ifx\\#1\\%
        \else
        \,#1%
        \fi
    }%
}
\newcommand*{\algoref}[2][]{%
    \hyperref[{algo:#2}]{%
        Algorithm~\ref*{algo:#2}%
        \ifx\\#1\\%
        \else
        \,#1%
        \fi
    }%
}
\newcommand*{\objref}[2][]{%
    \hyperref[{obj:#2}]{%
        Objective~\ref*{obj:#2}%
        \ifx\\#1\\%
        \else
        \,#1%
        \fi
    }%
}

\makeatletter
\newcommand{\printfnsymbol}[1]{%
  \textsuperscript{\@fnsymbol{#1}}%
}
\makeatother

\usepackage{xspace}
\makeatletter
\DeclareRobustCommand\onedot{\futurelet\@let@token\@onedot}
\def\@onedot{\ifx\@let@token.\else.\null\fi\xspace}

\def\eg{\emph{e.g}\onedot} \def\Eg{\emph{E.g}\onedot}
\def\ie{\emph{i.e}\onedot} 
 
 \def\vs{\emph{vs}\onedot}

\makeatother

\iccvfinalcopy % *** Uncomment this line for the final submission

\def\iccvPaperID{2556} % *** Enter the ICCV Paper ID here

\ificcvfinal\fi

\begin{document}

%%%%%%%%% TITLE
% \title{Improving Unmanned Aerial Vehicle-based Object Detection via Nuisance Disentanglement}
\title{Delving into Robust Object Detection from Unmanned Aerial Vehicles: \\ A Deep Nuisance Disentanglement Approach}

\author{
\vspace*{-1cm}
Zhenyu Wu$^{1}$\thanks{Work done during Z. Wu's internship at ARL},\,  Karthik Suresh$^{1}$, Priya Narayanan$^{2}$, Hongyu Xu$^{3}$\thanks{Currently works at Apple Inc.},\, Heesung Kwon$^{2}$, Zhangyang Wang$^{1}$
$^1$Texas A\&M University\hspace{1em} $^2$U.S. Army Research Laboratory\hspace{1em} $^3$University of Maryland
%\vspace*{-1cm}
}

\maketitle
\thispagestyle{empty}
%%%%%%%%% ABSTRACT
\begin{abstract}
%\vspace{-1em}
\vspace{-0.5em}
Object detection from images captured by Unmanned Aerial Vehicles (UAVs) is becoming increasingly useful. Despite the tremendous success of the generic object detection methods trained on ground-to-ground images, a considerable performance drop is observed when they are directly applied to images captured by UAVs. The unsatisfactory performance is owing to many UAV-specific nuisances, such as varying flying altitudes, adverse weather conditions, dynamically changing viewing angles, etc. Those nuisances constitute a large number of fine-grained domains, across which the detection model has to stay robust. Fortunately, UAVs will record meta-data that depict those varying attributes, which are either freely available along with the UAV images or can be easily obtained. We propose to utilize those free meta-data in conjunction with associated UAV images to learn domain-robust features via an adversarial training framework dubbed Nuisance Disentangled Feature Transform (\textbf{NDFT}), for the specific challenging problem of object detection in UAV images, achieving a substantial gain in robustness to those nuisances. We demonstrate the effectiveness of our proposed algorithm by showing the state-of-the-art performance (single model) on two existing UAV-based object detection benchmarks. The code is available at \url{https://github.com/VITA-Group/UAV-NDFT}.
\end{abstract}

\vspace{-1em}
\section{Introduction}
\vspace{-0.5em}
\begin{figure*}[t!]
\centering
\begin{tabular}{cc}
\subfloat[Baseline F-RCNN]{\includegraphics[width = 3.05in]{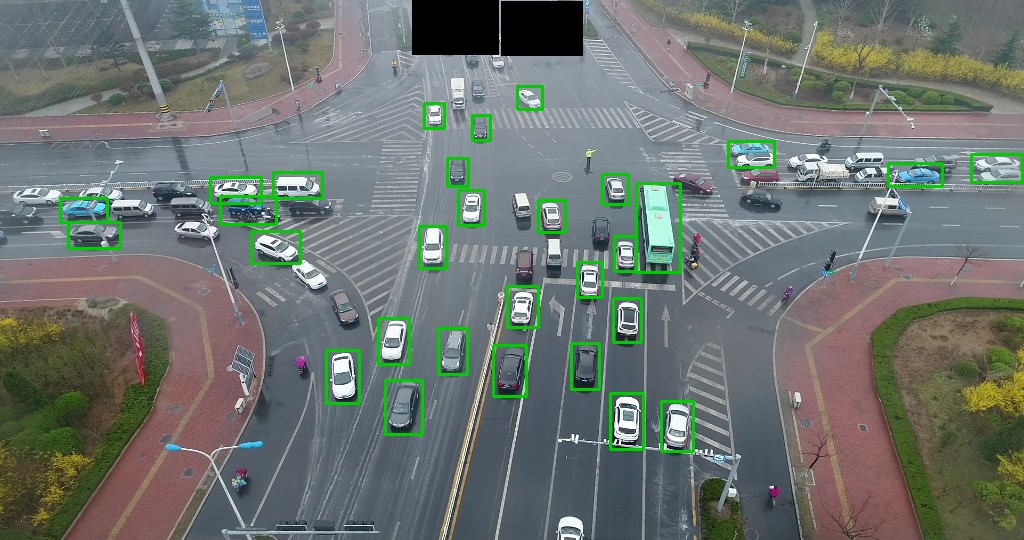}} &
\subfloat[NDFT-Faster-RCNN (A)]{\includegraphics[width = 3.05in]{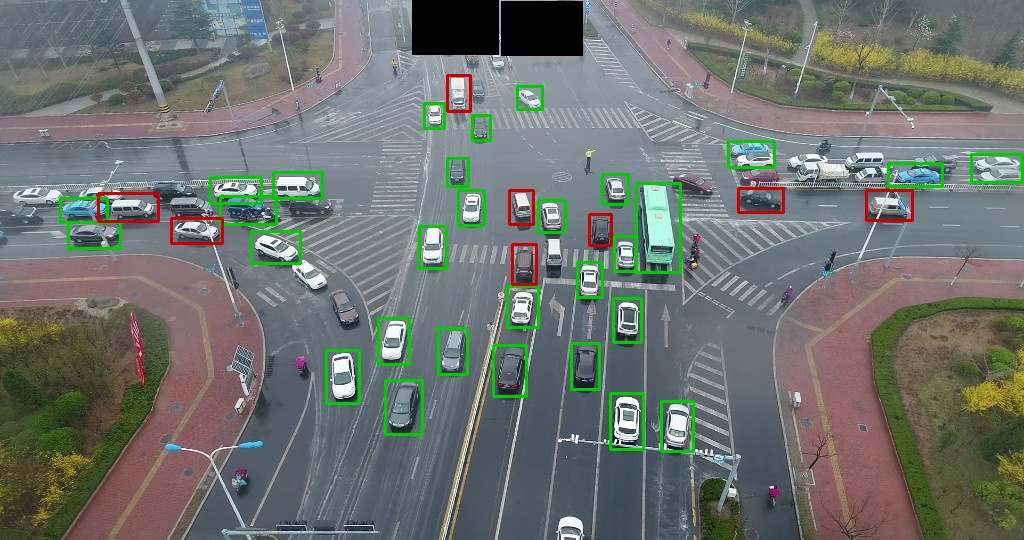}}\\
\subfloat[NDFT-Faster-RCNN (A+V)]{\includegraphics[width = 3.05in]{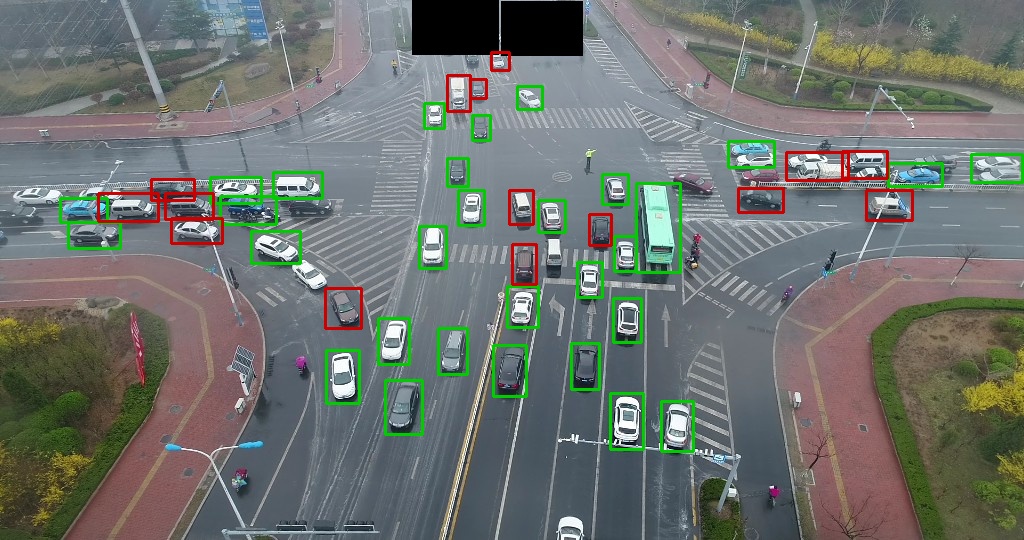}} &
\subfloat[NDFT-Faster-RCNN(A+V+W)]{\includegraphics[width =3.05in]{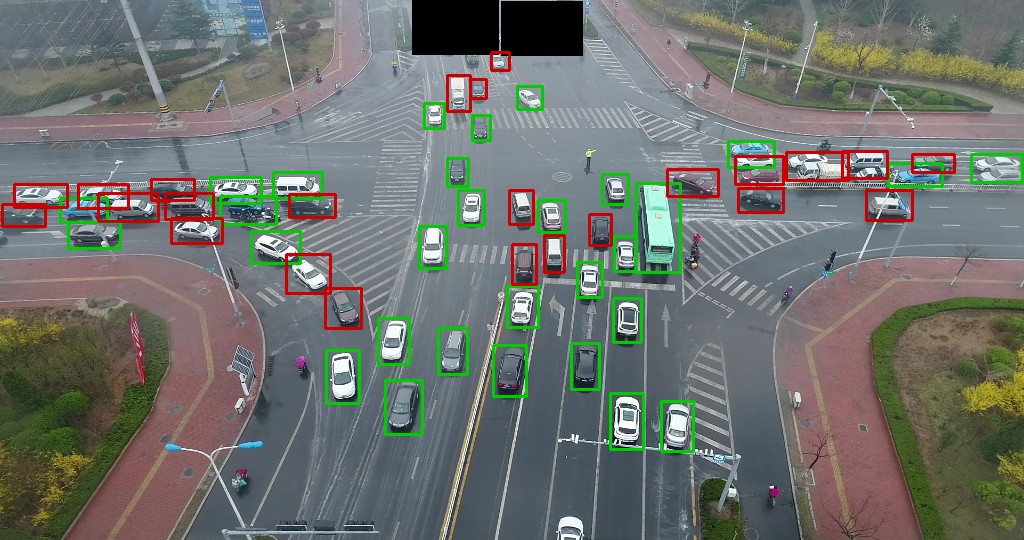}}\\
\end{tabular}
\vspace{-0.5em}
\caption{Examples showing the benefit of the proposed NDFT framework for object (vehicle) detection on the UAVDT dataset: starting from (a) Faster-RCNN~\cite{ren2015faster} baseline, to gradually (b) disentangling the nuisances of altitude (A); (c) disentangling the nuisances of both altitude (A) and view angles (V); and (d) disentangling all the nuisances of altitude (A), view angles (V), and weather (W). The detection performance gradually improves from (a) to (d) with disentanglement on more nuisances (red rectangular boxes denote new correct detections beyond the baseline).}
\vspace{-1em}
\end{figure*}

Object detection has been extensively studied over the decades. While most of the good detectors are able to detect objects of interest in clear images, such images are usually captured from ground-based cameras. With the rapid development of machinery technology, Unmanned Aerial Vehicles (UAVs) equipped with cameras have been increasingly deployed in many industrial applications, opening up a new frontier of computer vision applications in security surveillance, peacekeeping, agriculture, deliveries, aerial photography, disaster assistance \cite{semsch2009autonomous,honkavaara2013processing,DroneDelivery,erdelj2016uav,vidalmata2019bridging}, etc. One of the core features for the UAV-based applications is to detect objects of interest (\eg, pedestrians or vehicles). Despite high demands, object detection from UAV is yet insufficiently investigated. In the meantime, the large mobility of UAV-mounted cameras bring in greater challenges than traditional object detection (using surveillance or other ground-based cameras), such as but not limited to:
\begin{itemize}[leftmargin=*]
\vspace{-0.5em}
    \item \textbf{Variations in altitude and object scale}: The scales of objects captured in the image are closely affected by the flying altitude of UAVs. \Eg, the image captured by a DJI Inspire 2 series flying at 500 meters altitude \cite{DJIInspire2Spec} will contain very small objects, which are very challenging to detect and track.
    In addition, a UAV can be operated in a variety of altitudes while capturing images. When shooting at lower altitudes, its camera can capture more details of objects of interest. When it flies to higher altitudes, the camera can inspect a larger area, and more objects will be captured in the image. As a consequence, the same object can vary a lot in terms of scale throughout the captured video, with different flying altitudes during a single flight. 
    \vspace{-0.5em}
    \item \textbf{Variations in view angle}: The mobility of UAVs leads to video shoots from different and free angles, in addition to the varying altitudes. \Eg, a UAV can look at one object from the front view, to side view, to bird view, in a very short period of time. The diverse view angles cause arbitrary orientations and aspect ratios of the objects. Some view angles such as bird-view hardly occur in traditional ground-based object detection. As a result, the UAV-based detection model has to deal with more different visual appearances of the same object. Note that more view angles can be presented when altitudes grow higher. Also, wider view angles often lead to denser objects in the view.
    \vspace{-0.5em}
    \item \textbf{Variations in weather and illumination}: A UAV operated in uncontrolled outdoor environments may fly under various weather and lighting conditions. The changes in illumination (daytime versus nighttime) and weather conditions (\eg, sunny, cloudy, foggy, or rainy) will drastically affect the object visibility and appearance. 
    \vspace{-0.5em}
\end{itemize}

Most off-the-shelf detectors are trained with usually less varied, more restricted-view data. In comparison, the abundance of \textbf{UAV-specific nuisances} will cause the resulting UAV-based detection model to operate in a large number of different \textbf{fine-grained domains}. Here a domain could be interpreted as a specific combination of nuisances: \eg, the images taken at low-altitude and daytime, and those taken the high-altitude and nighttime domain, constitute two different domains. Therefore, our goal is to train a \textit{cross-domain object detection} model that stays robust to those massive number of fine-grained domains. Existing potential solutions include data augmentation \cite{DetectionAug,dvornik2018modeling}, domain adaption \cite{raj2015subspace,chen2018domain}, and ensemble of expert models \cite{lee2017me}. However, neither of these approaches are easy to generalize to multiple and/or unseen domains \cite{raj2015subspace,chen2018domain}, and they could lead to over-parameterized models which are not suitable for UAV on-board deployments \cite{DetectionAug,dvornik2018modeling,lee2017me}. 
%Discussion on relevant methods is detailed in Section 2.2.

\noindent \textbf{A (Almost) Free Lunch: Fine-Grained Nuisance Annotations}. In view of the above, we cast the UAV-based object detection problem as a cross-domain object detection problem with fine-grained domains. The object types of interest sustain across domains; such task-related features shall be preserved and extracted. The above UAV-specific nuisances constitute the domain-specific nuisances that should be eliminated for transferable feature learning. For UAVs, major nuisance types are well recognized, \eg, altitude, angle, and weather. More Importantly, in the specific case of UAVs, those nuisances annotations could be easily obtained or even freely available. \Eg, a UAV can record its flying altitudes as metadata by GPS, or more accurately, by a barometric sensor; weather information is easy to retrieve, since one can straightforwardly obtain the weather of a specific time/location with each UAV flight's time-stamp and spatial location (or path).

Motivated by those observations, we propose to learn an object detection model that maintains its effectiveness in extracting task-related features while eliminating the recognized types of nuisances across different domains (\eg, altitudes/angles/weathers). We take advantage of the free (or easy) access to the nuisance annotations. Based on them, we are the first to adopt an adversarial learning framework, to learn task-specific, domain-invariant features by explicitly disentangling task-specific and nuisance features in a supervised way. The framework, dubbed \textit{Nuisance Disentangled Feature Transform} (\textbf{NDFT}), gives rise to highly robust UAV-based object detection models that can be directly applicable to not only domains in training, but also more unseen domains, without needing any extra effort of domain adaptation or sampling/labeling. Experiments on two real UAV-based object detection benchmarks suggest the state-of-the-art effectiveness of NDFT.

\vspace{-0.5em}
\section{Related Works}
\vspace{-0.5em}
\subsection{Object Detection: General and UAV-Specific}
\vspace{-0.5em}
Object detection has progressed tremendously, partially thanks to established benchmarks (\ie, MS COCO~\cite{lin2014microsoft} and PASCAL VOC~\cite{everingham2010pascal}). There are primarily two main streams of approaches: two-stage detectors and single-stage detectors, based on whether the detectors have proposal-driven mechanism or not. Two stage detectors~\cite{girshick2014rich,he2014spatial,girshick2015fast,ren2015faster,dai2016r,xu2018deep,xu2019deep} contains region proposal network (RPN) to first generate region proposals, and then extract region-based features to predict the object categories and their corresponding locations. Single-stage detectors~\cite{redmon2016you,redmon2017yolo9000,redmon2018yolov3,liu2016ssd} apply dense sampling windows over object locations and scales, and usually achieved higher speed than two-stage ones, although often at the cost of (marginal) accuracy decrease.

\noindent\textbf{Aerial Image-based Object Detection}
A few aerial image datasets (\ie, DOTA~\cite{xia2018dota}, NWPU VHR-10~\cite{cheng2014multi}, and VEDAI~\cite{razakarivony2016vehicle} ) were proposed recently. However, those above datasets only contain geospatial images (\eg, satellite) with bird-view small objects, which are not as diverse as UAV-captured images with greatly more varied altitudes, poses, and weathers. Also, the common practice to detect objects from aerial images remains still to deploy off-the-shelf ground-based object detection models~\cite{han2016deep,narayanan2019overview}. 

Public benchmarks were unavailable for specifically UAV-based object detection until recently. Two datasets, UAVDT~\cite{du2018unmanned} and VisDrone2018~\cite{zhu2018vision}, were released to address this gap. UAVDT consists of 100 video sequences (about 80k frames) captured from UAVs under complex scenarios. Moreover, it also provides full annotations for weather conditions, flying altitudes, and camera views in addition to the ground truth bounding box of the target objects. VisDrone2018~\cite{zhu2018vision} is a large-scale UAV-based object detection and tracking benchmark, composed of 10,209 static images and 179,264 frames from 263 video clips.

\noindent\textbf{Detecting Tiny Objects} A typical ad-hoc approach to detect tiny objects is through learning representations of all the objects at multiple scales. This approach is, however, highly inefficient with limited performance gains. \cite{cao2016towards} proposed a super-resolution algorithm using coupled dictionary learning to transfer the target region into a high resolution to ``augment'' its visual appearance. \cite{wang2016studying,li2017perceptual,liu2019enhance} proposed to internally super-resolve the feature maps of small objects to make them resemble similar characteristics as large objects. SNIP~\cite{singh2018analysis} showed that CNNs were not naturally robust to the variations in object scales. It proposed to train and test detectors on the same scales of an image pyramid, and selectively back-propagate the gradients of object instances of different sizes as a function of the image scale during the training stage. SNIPER~\cite{singh2018sniper} further processed context regions around ground-truth instances at different appropriate scales to efficiently train the detector at multiple scales, improving the detection of tiny object detection more.

\vspace{-0.5em}
\subsection{Handling Domain Variances}
\vspace{-0.5em}
\noindent\textbf{Domain Adaptation via Adversarial Training}
Adversarial domain adaptation~\cite{ganin2014unsupervised} was proposed to reduce the domain gap by learning with only labeled data from a source domain plus massive unlabeled data from a target domain. This approach has recently gained increased attention in the detection fields too. \cite{wang2017fast} learned robust detection models to occlusion and deformations, through hard positive examples generated by an adversarial network. \cite{chen2018domain} improved the cross-domain robustness of object detection by enforcing adversarial domain adaption on both image and instance levels. \cite{bashmal2018siamese} introduced a Siamese-GAN to learn invariant feature representations for both labeled and unlabeled aerial images coming from two different domains. CyCADA~\cite{hoffman2017cycada} unified cycle-consistency with adversarial loss to learn domain-invariance. However, these domain adaption methods typically assume one (ideal) source domain and one (non-ideal) target domain. The possibility of generalizing these methodologies to handling many fine-grained domains is questionable. Once a new unseen domain emerges, domain adaptation needs explicit re-training. 

In comparison, our proposed framework does not assume any ideal reference (source) domain, but rather tries to extract invariant features shared by many different ``non-ideal'' target domains (both seen and unseen), by disentangling domain-specific nuisances. The setting thus differs from typical domain adaptation and generalizes to task-specific feature extraction in unseen domains naturally.

\noindent\textbf{Data Augmentation, and Model Ensemble}
Compared to the considerable amount of research in data augmentation for classification~\cite{ganin2014unsupervised}, less attention was paid to other tasks such as detection~\cite{DetectionAug}. Classical data augmentation relies on a limited set of pre-known factors (such as scaling, rotation, flipping) that are easy to invoke and adopt ad-hoc, minor perturbations that are unlikely to change labels, in order to gain robustness to those variations. However, UAV images will involve a much larger variety of nuisances, many of which are hard to ``synthesize'', \eg, images from different angles. \cite{dvornik2018modeling,zhang2018dada} proposed learning-based approaches to synthesize new training samples for detection. But they focused on re-combining foreground objects and background contexts, rather than re-composing specific nuisance attributes. Also, the (much) larger augmented dataset adds to the training burden and may cause over-parameterized models.
 
Another methodology was proposed in~\cite{lee2017me}. To capture the appearance variations caused by different shapes, poses, and viewing angles, it proposed a Multi-Expert R-CNN consisting of three experts, each responsible for objects with a particular shape: horizontally elongated, square-like, and vertically elongated. This approach has limitations as the model ensemble quickly becomes too expensive as more different domains are involved. It further cannot generalize to unknown or unseen domains. 

\noindent\textbf{Feature Disentanglement in Generative Models}
Feature disentanglement~\cite{xiang2017linear,uplavikar2019all} leads to non-overlapped groups of factorized latent representations, each of which would properly describe corresponding information to particular attributes of interest. It has mostly been applied to generative models~\cite{desjardins2012disentangling,siddharth2016learning}, in order to disentangle the factors of variation from the content in the latent feature space. In the image-to-image translation, a recent work~\cite{gonzalez2018image} disentangled image representations into shared parts for both domains and exclusive parts for either domain. NDFT extends the idea of feature disentanglement to learning cross-domain robust discriminative models. Due to the different application scope from generative models, we do not add back the disentangled components to reconstruct the original input.
\vspace{-0.5em}
\section{Our Approach}
\begin{figure*}
\centering{
  \includegraphics[width=0.9\textwidth]{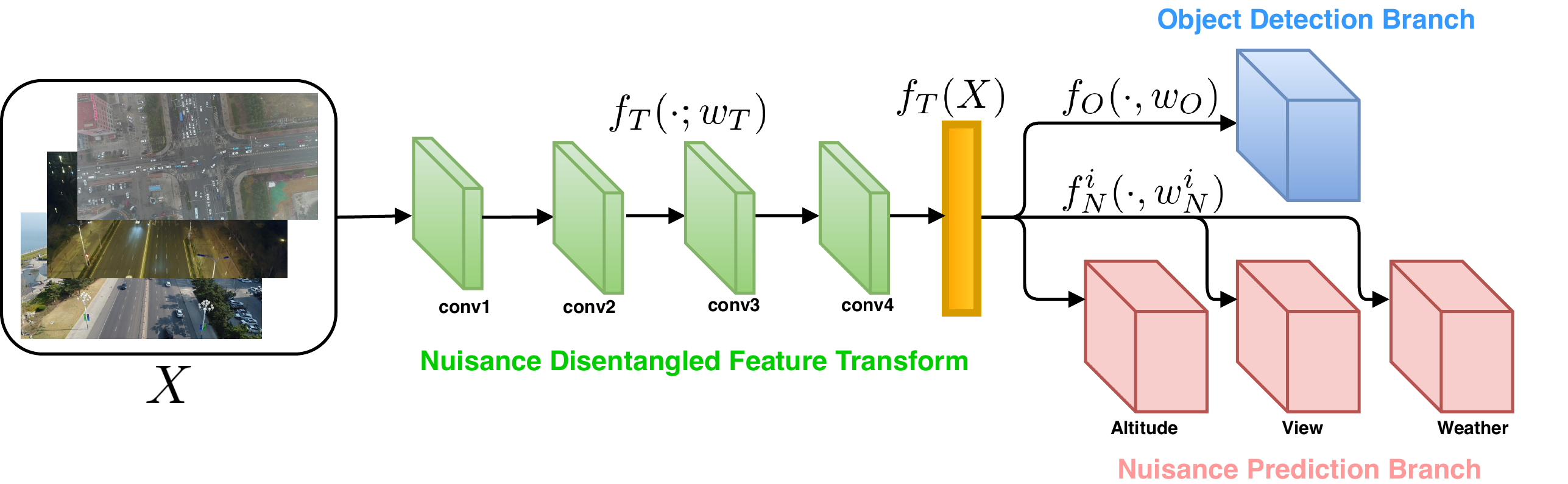}
  \vspace{-1.0em}
  \caption{Our proposed NDFT-Faster-RCNN network.}
  \vspace{-0.5em}
  \label{fig:NDFT-FRCNN}
}
\end{figure*}

\vspace{-0.5em}
\subsection{Formulation of NDFT}
\vspace{-0.5em}
Our proposed UAV-based cross-domain object detection can be characterized as an adversarial training framework. Assume our training data $X$ is associated with an \textbf{O}bject detection task $\mathcal{O}$, and a UAV-specific \textbf{N}uisance prediction task $\mathcal{N}$. 
%Apparently, the latter can be straightforwardly generated to multiple-nuisance cases via multi-label learning. 
We mathematically express the goal of cross-domain object detection as alternatively optimizing two objectives as follows ($\gamma$ is a weight coefficient): 
\vspace{-0.5em}
\begin{equation}
\begin{split}\label{obj:NDFT}
%\displaystyle\min_{f_O, f_T} \displaystyle\max_{f_N}L_O(f_O(f_{T}(X)), Y_O) - \gamma L_N (f_N(f_{T}(X)), Y_N)
\! \displaystyle\min_{f_O, f_T} & L_O(f_O(f_{T}(X)), Y_O) - \gamma L_N (f_N(f_{T}(X)), Y_N), \\
\displaystyle\min_{f_N}\ &L_N(f_N(f_{T}(X)), Y_N)
\vspace{-0.5em}
\end{split}
\end{equation}
In (\ref{obj:NDFT}), $f_O$ denotes the model that performs the object detection task $\mathcal{O}$ on its input data. The label set $Y_O$ are object bounding box coordinates and class labels provided on $X$. $L_O$ is a cost function defined to evaluate the object detection performance on $\mathcal{O}$. On the other hand, the labels of the UAV-specific nuisances $Y_N$ come from metadata along with $X$ (\eg, flying altitude, camera view or weather condition), and a standard cost function $L_N$ (\eg, softmax) is defined to evaluate the task performance on $\mathcal{N}$. Here we formulate nuisance robustness as the suppression of the nuisance prediction accuracy from the learned features.

We seek a \textit{Nuisance Disentangled Feature Transform} (\textbf{NDFT}) $f_T$ by solving (\ref{obj:NDFT}), such that 
\begin{itemize}
\vspace{-0.5em}
\item The object detection task performance $L_O$ is minimally affected over $f_T (X)$, compared to using $X$. 
\vspace{-0.5em}
\item The nuisance prediction task performance $L_N$ is maximally suppressed over $f_T (X)$, compared to using $X$.\vspace{-0.5em}
\end{itemize}
In order to deal with the multiple nuisances case, we extend the (\ref{obj:NDFT}) to multiple prediction tasks. Here we assume $k$ nuisances prediction tasks associated with label sets $Y_N^1,...,Y_N^{k}$. $\gamma_1,...,\gamma_{k}$ are the respective weight coefficients. The modified objective naturally becomes:
\vspace{-1em}
% \begin{equation}
% \begin{split}\label{obj:NDFT_multiple}
% \displaystyle\min_{f_O, f_T}\displaystyle\max_{f_N^1,...,f_N^k}&L_O(f_O(f_{T}(X)), Y_O) \\
% - \displaystyle\sum_{i=1}^{k}\gamma_i&L_N (f_N^i(f_{T}(X)), Y_N^i), \\
% \end{split}
% \end{equation} 
\begin{align}\label{obj:NDFT_multiple}
& \displaystyle\min_{f_O, f_T} L_O(f_O(f_{T}(X)), Y_O)-\displaystyle\sum_{i=1}^{k}\gamma_i L_N (f_N^i(f_{T}(X)), Y_N^i), \nonumber \\
& \displaystyle\min_{f_N^1,...,f_N^k} L_N(f_N^i(f_{T}(X)), Y_N^i)
\vspace{-3em}
\end{align}
%\textcolor{red}{If you use ``minus'' before the second term, why MAX??????}
$f_T$, $f_O$ and $f_N^i$s can all be implemented by deep networks. 

\noindent\textbf{Interpretation as Three-Party Game} NDFT can be derived from a \textit{three-competitor game} optimization:
\begin{eqnarray*}
\vspace{-2em}
\displaystyle\max_{f_N} \displaystyle\min_{f_O, f_T} & L_O(f_O(f_{T}(X)), Y_O) - \gamma L_N (f_N(f_{T}(X)), Y_N)
\vspace{-2em}
\end{eqnarray*}
where $f_T$ is an \textit{obfuscator}, $f_N$ as a \textit{attacker}, and $f_O$ as an \textit{utilizer} (adopting ML security terms). In fact, the two sub-optimizations in (\ref{obj:NDFT}) denote an iterative routine to solve this unified form (performing coordinate descent between \{$f_T$, $f_O$\}, and $f_N$). This form can easily capture many other settings or scenarios, \eg, privacy-preserving visual recognition \cite{wu2018towards,wang2019privacy} where $f_T$ encodes features to avoid peeps from $f_N$ while preserving utility for $f_O$. 
%We will revise Sec. 2 to include this theoretical ground of our approach. 

%\subsection{Problem Solution}

\begin{algorithm*}
\vspace{-0.2em}
\caption{Learning Nuisance Disentangled Feature Transform in UAV-based Object Detection via Adversarial Training}\label{algo:AdvAlg}
%\vspace{-0.5em}
\begin{algorithmic}
\State{Given pre-trained NDFT module $f_T$, object detection task module $f_O$, and nuisances prediction modules $f_N^i$s} 
\For{number of training iterations}
\State{Sample a mini-batch of {n} examples \{$X^1, \cdots, X^n$\}}
\State{Update \textbf{NDFT module} $f_T$ (weights $w_T$) and \textbf{object detection module} $f_O$ (weights $w_O$) with stochastic gradients:}
\State{ $\nabla_{w_T\cup w_O} \frac{1}{n}\displaystyle\sum_{j=1}^n \Big[L_O(f_O(f_T(X^j)), Y_O^j)+\displaystyle\sum_{i=1}^{k}\gamma_i L_{ne}(f_N^i(f_T(X^j)))\Big]$}
\While{at least one nuisance prediction task has training accuracy $\leq$ $0.9$} \Comment{Prevent $f_N^i$s from becoming too weak.}
%\State{Sample a new mini-batch of $n$ examples \{$X_1, \cdots, X_n$\}}
\State{Update \textbf{nuisance prediction modules} $f_N^i, \dots, f_N^k$ (weights $w_N^1,\dots,w_N^k$) with stochastic gradients:}
\State{$\nabla_{w_N^i}\frac{1}{n}\displaystyle\sum_{j=1}^n\displaystyle\sum_{i=1}^k{L_N(f_N^i(f_T(X^j)),Y_N^j)}$}
\EndWhile
\State {Restart $f_N^i, \dots, f_N^k$ every 1000 iterations, and repeat Algorithm 1 from the beginning.}\Comment{Alleviate overfitting.}
%\State %\Goto \texttt{marker}
%Repeat Algorithm 1 from the beginning.
\EndFor
\end{algorithmic}
\vspace{-0.2em}
\end{algorithm*}
\vspace{-0.5em}

%\vspace{-0.5em}
\subsection{Implementation and Training}
\vspace{-0.5em}
\noindent\textbf{Architecture Overview: NDFT-Faster-RCNN} As an instance of the general NDFT framework (\ref{obj:NDFT_multiple}), Figure \ref{fig:NDFT-FRCNN} displays an implementation example of NDFT using the Faster-RCNN backbone \cite{ren2015faster}, while later we will demonstrate that NDFT can be plug-and-play with other more sophisticated object detection networks (\eg, FPN). 

%We propose a Nuisance-Invariant Feature Transform Faster-RCNN in \figref{NDFT-FRCNN}. 
During training, the input data $X$ first goes through the NDFT module $f_T$, and its output $f_T(X)$ is passed through two subsequent branches simultaneously. The upper object detection branch $f_O$ uses $f_T(X)$ to detect objects, while the lower nuisance prediction model $f_N$ predicts nuisance labels from the same $f_T(X)$. Finally, the network minimizes the prediction penalty (error rate) for $f_T$, while maximizing the prediction penalty for $f_N$, shown by (\ref{obj:NDFT_multiple}).

By jointly training $f_T$, $f_O$, and $f_N^i$s in the above adversarial settings, the NDFT module will find the optimal transform that preserves the object detection related features while removing the UAV-specific nuisances prediction related features, fulfilling the goal of cross-domain object detection that is robust to the UAV-specific nuisances.

\noindent\textbf{Choices of $f_T$, $f_O$ and $f_N$} In this NDFT-Faster-RCNN example,  $f_T$ includes the conv1\_x, conv2\_x, conv3\_x and conv4\_x of the ResNet101 part of Faster-RCNN. 
%\textcolor{red}{be more specific: which faster rcnn version/what resnet you used??}. 
$f_O$ includes the conv5\_x layer, attached with a classification and regression loss for detection. We further implement $f_N$ using the same architecture as $f_O$ (except the number of classes for prediction). The output of $f_T$ is fed to $f_O$ after going through RoIAlign \cite{he2017mask} layer, while it is fed to $f_N$ after going through a spatial pyramid pooling layer \cite{he2014spatial}. 
%Differently, the output of $f_T$ goes through a spatial pyramid pooling layer \cite{he2014spatial} before being fed to $f_N$.
%\textcolor{red}{??? are you saying conv5 belong to both detection and nuisance branches??}. 
%...\textcolor{red}{how about feeding $f_O$? with SPP or not, why??}.
\noindent\textbf{Choices of $L_O$ and $L_N$} $L_O$ is the bounding box classification (\eg, softmax) and regression loss (\eg, smooth $\ell_1$) as widely used in traditional two-stage detectors. However, using $-L_N$ as the adversarial loss in the first row of (\ref{obj:NDFT_multiple}) is not straightforward. If $L_N$ is chosen as some typical classification loss such as the softmax, maximizing $L_N$ is prone to gradient explosion. 
After experimenting with several solutions such as the gradient reversal trick \cite{ganin2014unsupervised}, we decide to follow \cite{Liu_2018_CVPR} to choose the negative entropy function of the predicted class vector as the adversarial loss, denoted as $L_{ne}$. Minimizing $L_{ne}$ will encourage the model to make ``uncertain'' predictions (equivalently, close to uniform random guesses) on the nuisances.

Since we replace $L_N$ with $L_{ne}$ in the first objective in  (\ref{obj:NDFT_multiple}), it no longer needs $Y_N$. Meanwhile, the usage of $L_N$ and $Y_N$ remains unaffected in the second objective of (\ref{obj:NDFT_multiple}). $L_N$ and $Y_N$ are used to pre-train $f_N^i$s at the initialization and keep $f_N^i$s as ``sufficiently strong adversaries'' throughout the adversarial training, in order to learn meaningful $f_T$ that can generalize better. Our final framework alternates between:
\vspace{-1em}
\begin{align}\label{obj:NDFT_final}
&\displaystyle\min_{f_O, f_T} L_O(f_O(f_{T}(X)), Y_O) + \displaystyle\sum_{i=1}^{k}\gamma_i L_{ne} (f_{N}^i(f_{T}(X))), \nonumber \\
&\displaystyle \min_{f_N^1,...,f_N^k} L_N(f_N^i(f_{T}(X)), Y_N^i) 
\end{align}

% We also use $Y_N$ to pre-train strong $f_N^i$s at the initialization and maintain $f_N^i$s' performance during the adversarial training. 
%Furthermore, $Y_N$ will play a critical role in the domain budget model monitoring and restarting (represented by the second row of (\ref{obj:NDFT_multiple}))

% \textcolor{red}{Do you really have an ablation study on the three? I'm not positive to discuss too much loss options here, as it conflicts Ye' paper.}

% \textcolor{red}{Also, be consistent with your Algorithm 1!!}

% The first possible choice is the negative KL divergence between the predicted class vector and the ground truth label. It could be implemented as a gradient reversal layer (GRL)\cite{ganin2014unsupervised}. But minimizing a concave function will cause a ton of numerical instabilities (often explosions).

% The second possible choice is the KL divergence between the predicted class distribution and the uniform distribution.

% The third possible choice is the negative entropy function of the predicted class vector, which we found empirically works best among the three. Minimizing it will encourage the model to make ``uncertain'' predictions. Meanwhile, we will use $Y_N$ to ensure a sufficiently strong $f_N$ at the initialization. Furthermore, $Y_N$ will play a critical role in the domain budget model monitoring and restarting.

\noindent\textbf{Training Strategy}
Just like training GANs \cite{goodfellow2014generative}, our training is prone to collapse and/or bad local minima. We thus presented a carefully-designed training algorithm with the alternating update strategy. The training procedure is summarized in Algorithm \ref{algo:AdvAlg} and explained below.

For each mini-batch, we first jointly optimize $f_T$ and $f_O$ weights (with $f_N^i$s frozen), by minimizing the first objective in (\ref{obj:NDFT_final}) using the standard stochastic gradient descent (SGD). 
% following using the standard stochastic gradient descent solver:
% \begin{equation}
% \begin{split}\label{obj:step1}
% \displaystyle\min_{f_O, f_T} L_O(f_O(f_{T}(X)), Y_O) 
% + \displaystyle\sum_{i=1}^{k}\gamma_i L_{ne} (f_N^i(f_{T}(X))), \\
% \end{split}
% \end{equation} 
Meanwhile, we will keep ``monitering'' $f_N^i$ branches: as $f_T$ is updated, if at least one of the $f_N^i$ becomes too weak (\ie, showing poor predicting accuracy on the same mini-batch), another update will be triggered by minimizing the second objective in (\ref{obj:NDFT_final}) using SGD. 
% \begin{equation}
% \begin{split}\label{obj:step2}
% \displaystyle\min_{f_N^1,...,f_N^k} \sum_{i=1}^{k}\gamma_i L_N (f_N^i(f_{T}(X)), Y_N^i), \\
% \end{split}
% \end{equation} 
The goal is to ``strengthen'' the nuisance prediction competitors. 
%to keep them as ``sufficiently strong adversaries'' throughout training, in order to learn meaningful NDFT $f_T$ that can generalize better. 
Besides, we also discover an empirical trick, by periodically re-setting the current weights of $f_N^1,...,f_N^k$ to random initialization, and then re-train them on $f_T(X)$ (with $f_T$ fixed) to become strong nuisance predictors again, before we re-start the above alternative process of $f_T$, $f_O$ and $f_N^i$s. This re-starting trick is also found to benefit the generalization of learned $f_T$ \cite{wu2018towards}, potentially due to helping get out of some bad local minima.
\vspace{-1.5em}
\section{Experimental Results}
\vspace{-0.5em}

\begin{figure*}[t!]
\centering
\begin{tabular}{cc}
\subfloat[DE-FPN]{\includegraphics[width = 3.2in]{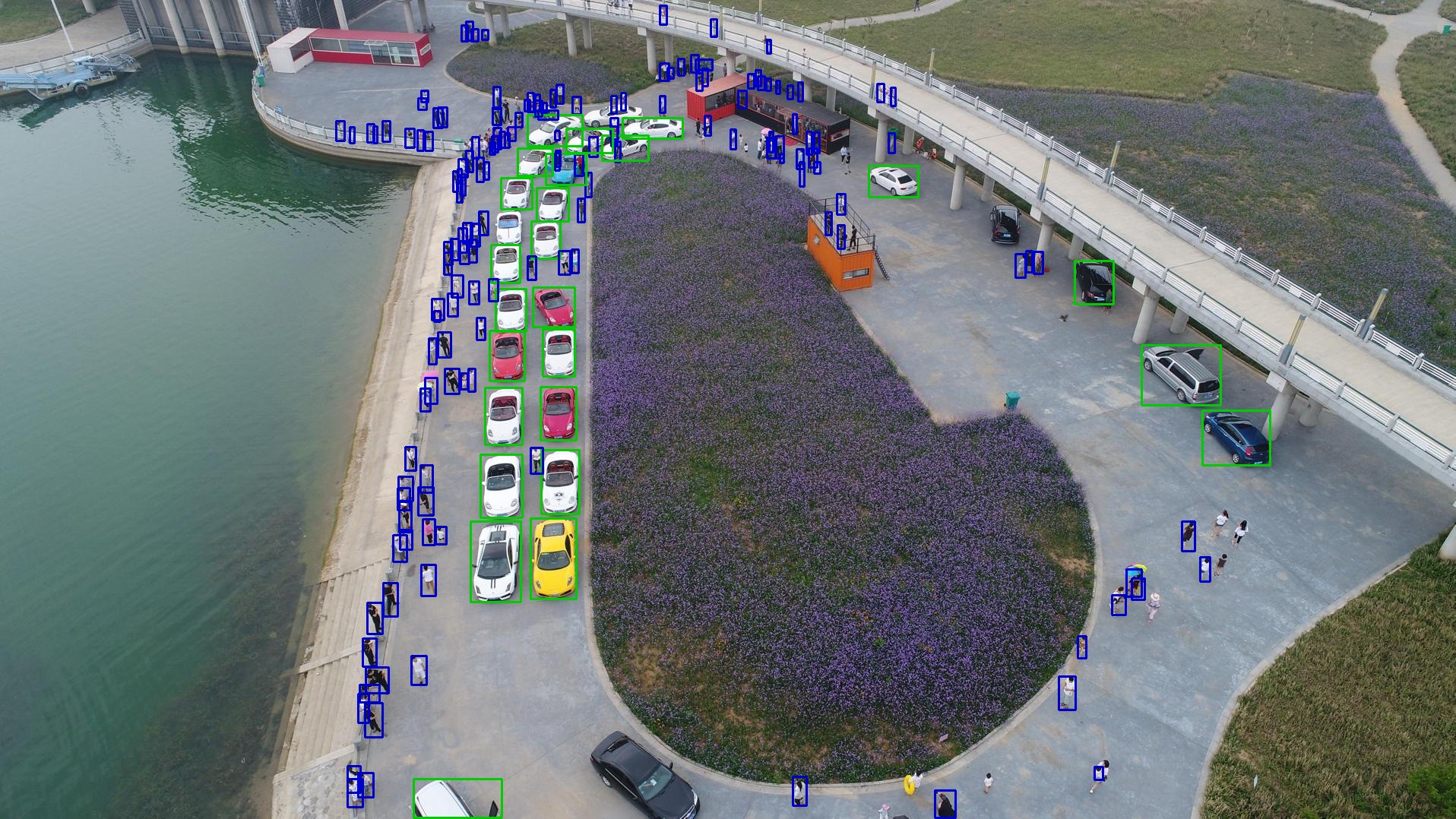}} &
\subfloat[NDFT-DE-FPN]{\includegraphics[width = 3.2in]{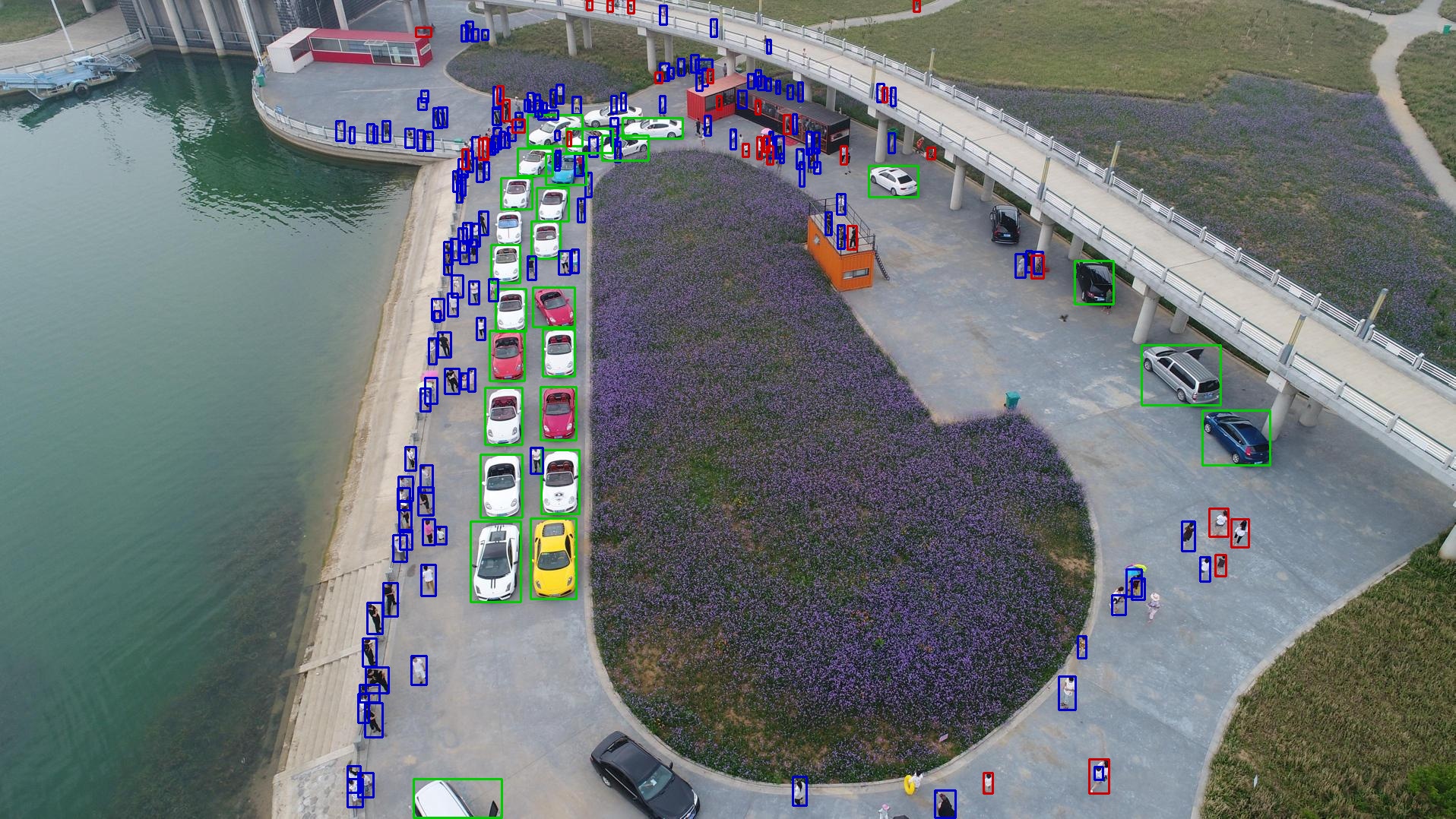}}\\
\end{tabular}
\vspace{-0.5em}
\caption{An example showing the benefit of the proposed NDFT approach for object detection on VisDrone2018 dataset. The blue and green rectangular boxes denote pedestrians and cars respectively. Red rectangular boxes denote new correctly detected objects by NDFT-DE-FPN beyond the baseline of DE-FPN.}
\label{visdrone_vis}
\vspace{-1em}
\end{figure*}
Since public UAV-based object detection datasets (in particular those with nuisance annotations) are currently of very limited availability, we design \textbf{three sets of experiments} to validate the effectiveness, robustness, and generality of NDFT. First, we  perform the main body of experiments on the \textbf{UAVDT} benchmark \cite{du2018unmanned}, which provides all three UAV-specific nuisance annotations (altitude, weather, and view angle). We demonstrate the clear observation that the more variations are disentangled via NDFT, the larger AP improvement we will gain on UAVDT; and eventually we achieve the state-of-the-art performance on UAVDT. 

We then move to the other public benchmark, \textbf{VisDrone2018}. Originally, the nuisance annotations were not released on VisDrone2018. We manually annotate the nuisances on each image: those annotations will be released publicly, and hopefully will be contributed as a part of VisDrone. Learning NDFT gives a performance boost over the best single model, and leads us to the (single model) state-of-the-art mean average precision (mAP)\footnote{mAP on the 10 categories of objects is the standard evaluation criterion on VisDrone2018.} on VisDrone2018 validation set\footnote{The top-2 models on the UAVDT leaderboard are model \textit{ensembles}. We compare with only \textit{single model} solutions for fairness.}.

In addition, we study a \textit{transfer learning} setting from the NDFT learned on UAVDT, to VisDrone2018. The goal of exploring transfer is because UAVs often come across unseen scenarios, and a good transferability of learned features facilitates general usability. When detecting the (shared) vehicles category, $f_T$ shows strong transferability by outperforming the best single-model method currently reported on the VisDrone2018 leaderboard \cite{VisDroneLeaderboard}.

\subsection{UAVDT: Results and Ablation Study}
\vspace{-0.5em}
\noindent\textbf{Problem Setting}
The image object detection track on UAVDT consists of around 41k frames with 840k bounding boxes. It has three categories: car, truck, and bus, but the class distribution is highly imbalanced (the latter two occupy less than 5\% of bounding boxes). Hence following the convention by the authors in  \cite{du2018unmanned}, we combine the three into one \textit{vehicle} class and report AP based on that. All frames are also annotated with three categories of UAV-specific nuisances: flying altitude (\textit{low, medium, and high}), camera views (\textit{front-view, side-view, and bird-view}), and weather condition\footnote{We discard another ``foggy'' class because of its too small size.}(\textit{daylight, night}). We will denote the three nuisances as \textbf{A}, \textbf{V}, and \textbf{W} for short, respectively. 

\noindent\textbf{Implementation Details}
We first did our best due diligence to improve the baseline (without considering nuisance handling) on UAVDT, to ensure a solid enough ground for NDFT. The authors reported an AP of $\sim$20 using a Faster-RCNN model with the VGG-16 backbone. We replace the backbone with ResNet-101, and fine-tune hyperparameters such as anchor scale (16,32,64,128,256). We end up with an improved AP of 45.64 (using the same IoU threshold = 0.7 as the authors) as our baseline performance. We also communicated with the authors of \cite{du2018unmanned} in person, and they acknowledged this improved baseline. We then implement NDFT-Faster-RCNN using the architecture depicted in Figure \ref{fig:NDFT-FRCNN}, also with a ResNet-101 backbone. We denote $\gamma_1$, $\gamma_2$ and $\gamma_3$ as the coefficients in (\ref{UAVDT_alt}), for the $L_{ne}$ loss terms for altitude, view and weather nuisances, respectively. 

\begin{table*}[!htb]
\footnotesize
\begin{minipage}[t]{.34\linewidth}
\caption{Learning NDFT-Faster-RCNN on altitude nuisance only, with different $\gamma_1$ values on the UAVDT dataset.} % title of Table
\centering % used for centering table
\begin{tabular}{c|c|c|c|c} % centered columns (4 columns)
\hline
\backslashbox{$\gamma_1$}{A} & Low & Med & High & Overall \\
\hline
0.0 & 68.14 & 49.71 & \textbf{18.70} & 45.64 \\
\hline
0.01 & \textbf{69.01} & 50.46 & 14.63 & 45.31 \\
\hline
0.02  & 66.97 & 46.91 & 16.69 & 44.17 \\ 
\hline
0.03  & 66.38 & \textbf{53.00} & 15.69 & \textbf{45.92} \\ 
\hline
0.05 & 65.46 & 48.43 & 16.58 & 44.36 \\ 
\hline
\end{tabular}
\label{UAVDT_alt}
\end{minipage}%
\hfill%
\begin{minipage}[t]{.34\linewidth}
\caption{Learning NDFT-Faster-RCNN on view angle nuisance only, with different $\gamma_2$ values on the UAVDT dataset.}
\centering
\begin{tabular}{c|c|c|c|c} % centered columns (4 columns)
\hline
\backslashbox{$\gamma_2$}{V} & Front & Side & Bird & Overall \\
\hline
0.0 & 53.34 & 68.02 & \textbf{27.05} & 45.64 \\
\hline
0.01 & 57.45 & 67.61 & 25.60 & \textbf{46.16} \\
\hline
0.02 & 61.49 & 66.85 & 24.93 & 45.73 \\ 
\hline
0.03 & 54.55 & \textbf{68.22} & 23.07 & 45.42 \\
\hline
0.04 & \textbf{64.93} & 66.83 & 24.96 & 46.10 \\ 
\hline
\end{tabular}
\label{UAVDT_angle}
\end{minipage}%
\hfill%
\begin{minipage}[t]{.28\linewidth}
\caption{Learning NDFT-Faster-RCNN on weather nuisance only, with different $\gamma_3$ values} % title of Table
\centering % used for centering table
\begin{tabular}{c|c|c|c} % centered columns (4 columns)
\hline
\backslashbox{$\gamma_3$}{W} & Day & Night & Overall \\
\hline
0.0 & \textbf{45.63} & 52.14 & 45.64 \\
\hline
0.01 & 45.18 & \textbf{59.66} & \textbf{46.62} \\
\hline
0.025 & 43.72 & 57.41 & 44.43 \\ 
\hline
0.05 & 43.89 & 50.25 & 43.79 \\
\hline
0.1 & 44.28 & 48.78 & 43.60 \\ 
\hline
\end{tabular}
\label{UAVDT_weather} % is used to refer this table in the text
\end{minipage}
\vspace{-2em}
\end{table*}

\noindent\textbf{Results and Analysis}
We unfold our full ablation study on UAVDT in a progressive way: first we study the impact of removing each individual nuisance type (A, V, and W). We then gradually proceed to remove two and three nuisance types and show the resulting consistent gains.

Tables \ref{UAVDT_alt}, \ref{UAVDT_angle}, and  \ref{UAVDT_weather} show the benefit of removing flying altitude (A), camera view (V) and weather condition (W) nuisances, individually. That could be viewed as learning NDFT-Faster-CNN (Figure \ref{fig:NDFT-FRCNN}) with only the corresponding one $\gamma_i$ ($i$ = 1, 2, 3) to be nonzero. The baseline model without nuisance disentanglement has $\gamma_i$  = 0, $i$ = 1, 2, 3. 

As can be seen from Table \ref{UAVDT_alt}, compared to the baseline ($\gamma_1$ = 0), an overall AP gain is obtained at $\gamma_1=0.03$, where we achieve a AP improvement of 0.28.

Table \ref{UAVDT_angle} shows the performance gain by removing the camera view (V) nuisance. At $\gamma_2=0.01$, an overall AP improvement of 0.52 is obtained. Similar positive observations are found in Table \ref{UAVDT_weather} as well, when the weather (W) nuisance is removed: $\gamma_3=0.01$ results in an overall AP boost of 0.98 over the baseline, with the more challenging night class AP increased by 7.52.

Table \ref{combined-tbd} shows the full results by incrementally adding more adversarial losses into training. For example, $A+V+W$ stands for simultaneously disentangling flying altitude, camera view, and weather nuisances. When using two or three losses, unless otherwise stated, we apply $\gamma_i$ = 0.01 for both/all of them, as discovered to give the best single-nuisance results in Tables \ref{UAVDT_alt} - \ref{UAVDT_weather}. 
As a consistent observation throughout the table, the more nuisances removed through NDFT, the better AP values we obtain (\eg, $A+V$ outperforms any of the three single models, and $A+V+W$ further achieves the best AP among all). In conclusion, removing nuisances using NDFT evidently contributes to addressing the tough problem of object detection on high-mobility UAV platforms. Furthermore, the final best-performer $A+V+W$ improves the class-wise APs noticeably on some most challenging nuisance classes, such as high-altitude, bird-view, and nighttime. Improving object detection in those cases can be significant for deploying camera-mounted UAVs to uncontrolled, potentially adverse visual environments with better reliability and robustness. 
\vspace{-0.5em}
\begin{table}[htb!]
\small
\caption{UAVDT NDFT-Faster-RCNN with multiple attribute disentanglement.} % title of Table
\vspace{-0.5em}
\centering % used for centering table
\resizebox{\columnwidth}{!}{
\begin{tabular}{c|c|c|c|c|c|c|c|c} 
\hline
{} & Baseline & A & V & W & A+V & A+W & V+W & A+V+W\\ 
\hline
& \multicolumn{8}{c}{Flying Altitude}\\
\hline
Low & 68.14 & 66.38 & 71.09 & \textbf{75.32} & 66.05 & 68.61 & 66.89 & 74.84 \\ 
\hline
Med & 49.71 & 53.00 & 52.29 & 51.59 & 54.07 & 49.18 & 56.07 & \textbf{56.24} \\
\hline
High & 18.70 & 15.69 & 16.62 & 16.08 & 18.60 & 19.19 & 15.42 & \textbf{20.55}\\ 
\hline
& \multicolumn{8}{c}{Camera View}\\
\hline
Front & 53.34 & 53.90 & 57.45 & 62.36 & 61.23 & 51.05 & 56.67 & \textbf{64.88} \\ 
\hline
Side & 68.02 & 67.41 &  67.61 & 68.47 & \textbf{68.82} & 68.71 & 67.62 & 67.50 \\ 
\hline
Bird & 27.05 & 24.56 & 25.60 & 23.97 & 24.43 & 27.96 & 24.41 & \textbf{28.79} \\
\hline
& \multicolumn{8}{c}{Weather Condition} \\
\hline
Day & 45.63 & \textbf{47.32} & 45.30 & 45.18 & 46.26 & 45.19 & 45.90 & 45.91 \\
\hline
Night & 52.14 & 45.82 & 56.70 & 59.66 & 59.16 & 59.78 & 53.35& \textbf{64.16} \\
\hline
\hline
Overall & 45.64 & 45.92 & 46.16 & 46.62 & 46.88 & 46.64 & 46.03 & \textbf{47.91} \\
\hline
\end{tabular}}
\label{combined-tbd} % is used to refer this table in the text
\vspace{-1.5em}
\end{table}

\noindent\textbf{Adopting Stronger FPN Backbones} We demonstrate that the performance gain by NDFT does not vanish as we adopt more sophisticated backbones, \eg, FPN \cite{lin2017feature}. Training FPN on UAVDT leads to the baseline performance improved from 45.64 to 49.05. By replacing Faster-RCNN with FPN in the NDFT training pipeline, the resulting model learns to simultaneously disentangle $A+V+W$ nuisances ($\gamma_i$ = 0.005, $i$ = 1,2,3). We are able to further increase the overall AP to 52.03, showing the general benefit of NDFT regardless of the backbone choices.

\vspace{0.4em}

\noindent\textbf{Proof-of-Concepts for NDFT-based Tracking } With object detection as our main focus, we also evaluate NDFT on UAVDT tracking for proof-of-concept. SORT \cite{bewley2016simple} (a popular online and real-time tracker) is chosen and evaluated on the multi-object tracking (MOT) task defined on UAVDT. We follow the tracking-by-detection framework adopted in \cite{du2018unmanned}, and compare the tracking results based on the detection inputs from vanilla Faster-RCNN and NDFT-Faster-RCNN ($A+V+W$), respectively. All evaluation protocols are inherited from \cite{du2018unmanned}. As in Table \ref{tracking}, NDFT-FRCNN largely outperforms the vanilla baseline in 10 out of the 11 metrics, showing its promise even beyond detection.

\begin{table}[h!]
\vspace{-0.5em}
\caption{NDFT \vs vanilla baseline on MOT task.} % title of Table
\centering % used for centering table
\vspace{-0.5em}
\resizebox{\columnwidth}{!}{
\begin{tabular}{c|ccc|cccc|cccc} 
\hline
{} & IDF & IDP & IDR & MOTA & MOTP & MT[\%] & ML[\%] & FP & FN & IDS & FM \\ 
\hline
FRCNN & 43.7 & 58.9 & 34.8 & \textbf{39.0} & 74.3 & 33.9 & 28.0 & 33,037 & 172,628 & 2,350 & 5,787\\
\hline
NDFT-FRCNN & \textbf{52.9} & \textbf{66.8} & \textbf{44.5} & 38.4 & \textbf{76.5} & \textbf{39.8} & \textbf{27.3} & \textbf{32,581} & \textbf{152,379} & \textbf{1,550} & \textbf{5,026}\\
\hline\end{tabular}}
\label{tracking} % is used to refer this table in the text
\vspace{-1em}
\end{table}

\begin{figure*}[h!]
\centering
\begin{tabular}{cc}
\vspace{-1em}
\subfloat[DE-FPN]{\includegraphics[width = 3in]{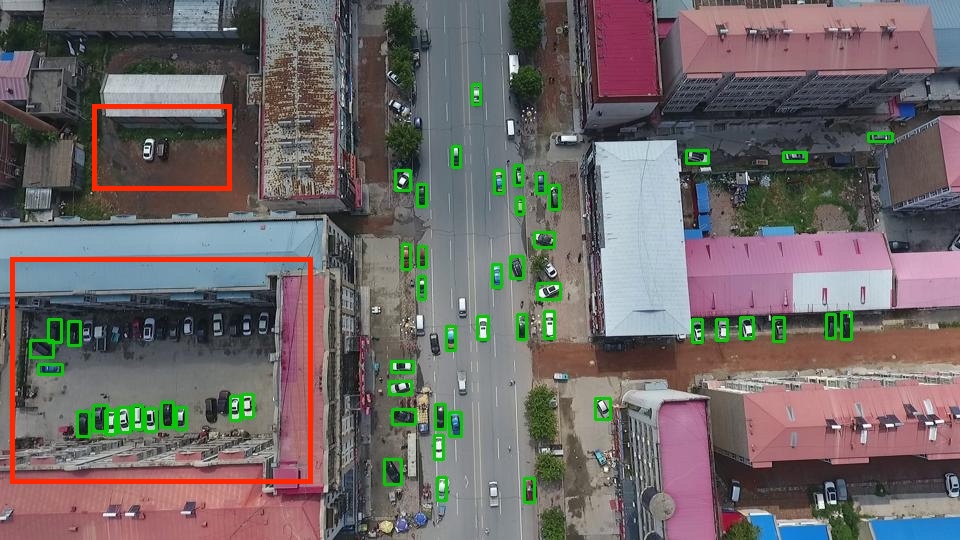}} &
\subfloat[NDFT-DE-FPN(r)]{\includegraphics[width = 3in]{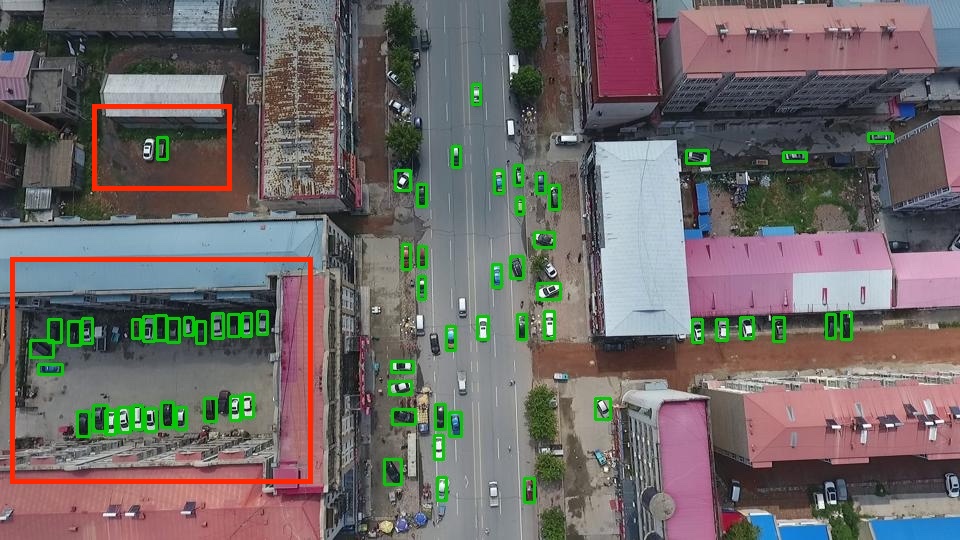}}\\
\end{tabular}
%\vspace{-0.5em}
\caption{An example showing the superior performance of NDFT-DE-FPN(r) over DE-FPN for object detection on VisDrone2018 dataset. Red boxes highlight the local regions where NDFT-DE-FPN(r) is able to detect substantially more vehicles than DE-FPN (the state-of-the-art single-model method on VisDrone2018). }
\vspace{-2.0em}
\label{visdrone_transfer_vis}
\end{figure*}
%\vspace{-1.0em}

\noindent\textbf{Comparing NDFT with Multi-Task Learning}
Another plausible option to utilize nuisance annotations is to jointly predict $Y_O$ and $Y_N^i$s as standard multi-task learning. To compare it with NDFT fairly, we switch the sign from $-$ to $+$ in (\ref{obj:NDFT_multiple}) first row, through which the nuisance prediction tasks become three auxiliary losses (\textbf{AL}) in multi-task learning. We minimize this new optimization and carefully re-tune $\gamma_i$s for AL by performing a grid search. As seen from Table \ref{mkl}, while AL is able to slightly improve over the baseline too (as expected), NDFT is evidently and consistently better thanks to its unique ability to encode invariances. The experiments objectively establish the role of adversarial losses versus standard auxiliary losses.

\begin{table}[h!]
\small
\vspace{-0.5em}
\caption{Comparing the baseline Faster-RCNN, adding auxiliary losses, and our proposed NDFT method.} % title of Table
\vspace{-0.5em}
\centering % used for centering table
\resizebox{\columnwidth}{!}{
\begin{tabular}{c|c|c|c|c|c|c|c|c|c} 
\hline
& & \multicolumn{3}{c|}{Altitude} & \multicolumn{3}{c|}{View} & \multicolumn{2}{c}{Weather} \\
\hline
{} & Overall  & Low & Med & High & Front & Side & Bird & Day & Night\\
\hline
Baseline & 45.64 & 68.14 & 49.71 & 18.70 & 53.34 & 68.02 & 27.05 & 45.63 & 52.14\\
\hline
AL & 45.69 & 66.58 & 50.80 & 18.28 & 61.49 & 66.85 & 24.93 & 45.62 & 53.64\\
\hline
NDFT & 46.81  & 70.48 & 55.06 & 16.12 & 57.06 & 68.07 & 27.59 & 46.05 & 59.56 \\
\hline
\end{tabular}}
\label{mkl} % is used to refer this table in the text
\vspace{-1em}
\end{table}

\subsection{VisDrone2018: Results and Analysis}
\paragraph{Problem Setting}
The image object detection track on VisDrone2018 provides a dataset of 10,209 images, with 10 categories of pedestrians, vehicles, and other traffic objects annotated. We manually annotate the UAV-specific nuisances, with the same three categories as on UAVDT.

According to the leaderboard \cite{VisDroneLeaderboard} and workshop report \cite{VisDrone2018DET}, the best-performing single model is DE-FPN, which utilized FPN (removing P6) with a ResNeXt-101 64-4d backbone. We implement DE-FPN by identically following their method description in \cite{VisDrone2018DET}, as our comparison subject.

\vspace{-1em}
\paragraph{Implementation Details}
Taking the DE-FPN backbone, NDFT is learned by simultaneously disentangling three nuisances (A+V+W). We create the DE-FPN model with NDFT, termed as NDFT-DE-FPN. 
The performance of DE-FPN and NDFT-DE-FPN are evaluated using the mAP over the 10 object categories on the VisDrone2018 validation set since the testing set is not publicly accessible.
\vspace{-1em}
\begin{table}[H]
\begin{center}
\caption{mAP comparison on VisDrone2018 validation set.} 
\vspace{-1em}% title of Table
\centering % used for centering table
\resizebox{\columnwidth}{!}{
\begin{tabular}{c|c|c|c|c|c|c|c} % centered columns (4 columns)
\hline 
{} &DE-FPN & \multicolumn{6}{c}{NDFT-DE-FPN} \\
\hline
$\gamma_i$ ($i$ = 1,2,3) & 0 & 0.001 & 0.003 & 0.004 & 0.005 & 0.01 & 0.02\\ 
\hline
mAP & 48.41 & 48.97 & 49.75 & 51.66 & \textbf{52.77} & 51.67 & 50.42 \\
\hline
\end{tabular}
}
\label{visdrone_results} % is used to refer this table in the text
\end{center}
\vspace{-2.5em}
\end{table}
\paragraph{Results and Analysis}
As in Table \ref{visdrone_results}, NDFT-DE-FPN gives rise to a 4.36 mAP boost over DE-FPN, making it a new state-of-the-art single model on VisDrone2018. Figure \ref{visdrone_vis} shows a visual comparison example.

%\vspace{-0.5em}
\subsection{Transfer from UAVDT to VisDrone2018}
\paragraph{Problem Setting}
We use VisDrone2018 as a testbed to showcase the transferablity of NDFT features learned from UAVDT. We choose DE-FPN as the comparison subject.

\vspace{-1em}
\paragraph{Implementation Details}
DE-FPN is trained on VisDrone 2018 training set and tested on the vehicle category of the validation set.
We then train the same DE-FPN backbone on UAVDT with three nuisances (A+V+W) disentangled ($\gamma_1 = \gamma_2 = \gamma_3 = 0.005$). The learned $f_T$ is then transferred to VisDrone2018, by only re-training the classification/regression layer while keeping other featured extraction layers all fixed. In that way, we focus on assessing the learned feature transferability using NDFT. Besides, we repeat the same above routine with $\gamma_1 = \gamma_2 = \gamma_3 = 0$, to create a transferred DE-FPN baseline without nuisance disentanglement. We denote the two transferred models as NDFT-DE-FPN(r) and DE-FPN(r), respectively. Since the vehicle is the only shared category between UAVDT and VisDrone2018, we compare average precision on the vehicle class only to ensure a fair transfer setting. The performance of DE-FPN, NDFT-DE-FPN(r), and DE-FPN(r) are compared on the VisDrone 2018 validation set (since the testing set is not publicly accessible).

\vspace{-1.5em}
\paragraph{Results and Analysis}
 The APs of DE-FPN, DE-FPN(r) and NDFT-DE-FPN(r) are 76.80, 75.27 and 79.50, receptively on the vehicle category.

Directly transferring DE-FPN from UAVDT to VisDrone2018 (fine-tuned on the latter) does not give rise to competitive performance, showing a substantial domain mismatch between the two datasets. 

However, transferring the learned NDFT to VisDrone2018 leads to performance boosts, with a 4.23 AP margin over the transfer baseline without disentanglement, and 2.70 over DE-FPN. It demonstrates that NDFT could potentially contribute to a more generally transferable UAV object detector that handles more unseen scenes (domains). A visual comparison example on VisDrone2018 is presented in Figure \ref{visdrone_transfer_vis}.

\vspace{-0.7em}
\section{Conclusion}
\vspace{-0.5em}
This paper investigates object detection from UAV-mounted cameras, a vastly useful yet under-studied problem. The problem appears to be more challenging than standard object detection, due to many UAV-specific nuisances. We propose to gain robustness to those nuisances by explicitly learning a Nuisance Disentangled Feature Transform (NDFT), utilizing the ``free'' metadata. Extensive results on real UAV imagery endorse its effectiveness. 

% {\small
% \bibliographystyle{ieee_fullname}
% \bibliography{egbib}
% }
\end{document}